\documentclass[sigconf]{acmart}
\usepackage{todonotes}
\usepackage{graphicx}
\usepackage{subcaption}
\usepackage[symbol]{footmisc}
\usepackage[ruled,vlined]{algorithm2e}

\AtBeginDocument{%
  \providecommand\BibTeX{{%
    \normalfont B\kern-0.5em{\scshape i\kern-0.25em b}\kern-0.8em\TeX}}}

\setcopyright{acmcopyright}
\copyrightyear{2020}
\acmYear{2020}
\acmDOI{10.1145/1122445.1122456}

\acmConference[FDG '20]{FDG '20: Foundations of Digital Games}{September 15--18, 2020}{Malta}
\acmBooktitle{FDG '20: Foundations of Digital Games,
  September 15--18, 2020, Malta}
\acmPrice{15.00}
\acmISBN{978-1-4503-XXXX-X/18/06}

\begin{document}

\title{Mario Level Generation From Mechanics Using Scene Stitching}



\author{Michael Cerny Green}
\authornote{Both authors contributed equally to this research.}
\affiliation{%
  \institution{New York University}
  \streetaddress{370 Jay Street}
  \city{Brooklyn}
  \state{NY}
  \postcode{11201}
}
\email{mike.green@nyu.edu}

\author{Luvneesh Mugrai}
\authornotemark[1]
\affiliation{%
  \institution{New York University}
  \streetaddress{370 Jay Street}
  \city{Brooklyn}
  \state{NY}
  \postcode{11201}
}
\email{lm3300@nyu.edu}

\author{Ahmed Khalifa}
\affiliation{%
  \institution{New York University}
  \streetaddress{370 Jay Street}
  \city{Brooklyn}
  \state{NY}
  \postcode{11201}
}
\email{ahmed@akhalifa.com}

\author{Julian Togelius}
\affiliation{%
  \institution{New York University}
  \streetaddress{370 Jay Street}
  \city{Brooklyn}
  \state{NY}
  \postcode{11201}
}
\email{julian@togelius.com}

\renewcommand{\shortauthors}{Green \& Mugrai et al.}

\begin{abstract}
This paper presents a level generation method for Super Mario by stitching together pre-generated ``scenes'' that contain specific mechanics, using mechanic-sequences from agent playthroughs as input specifications. Given a sequence of mechanics, our system uses an FI-2Pop algorithm and a corpus of scenes to perform automated level authoring. The system outputs levels that have a similar mechanical sequence to the target mechanic sequence but with a different playthrough experience. We compare our system to a greedy method that selects scenes that maximize the target mechanics. Our system is able to maximize the number of matched mechanics while reducing emergent mechanics using the stitching process compared to the greedy approach.
\end{abstract}



\maketitle

\section{Introduction}
Procedural Content Generation (PCG) methods are used to create content for games and simulations algorithmically. A large variety of PCG techniques exist, which have been applied a wide variety of types of game content and styles. Sometimes developers write down the rules that a program must abide to when generating, like a constraint-satisfaction generative system~\cite{smith2011answer}. Other methods let users define objective functions to be optimized by for example evolutionary algorithms~\cite{togelius2011search}, and yet others rely on supervised learning~\cite{summerville2018procedural}, or reinforcement learning~\cite{khalifa2020pcgrl}. 
Experience Driven PCG~\cite{yannakakis2011experience} is an approach to PCG that generates content so as to attempt to enable specific player experiences. This includes attempts to produce stylized generated content such as levels of specific difficulty~\cite{ashlock2010automatic,ashlock2011search}, for different playstyles~\cite{khalifa2018talakat,shaker20112010}, or even for a specific player's playing ability~\cite{yannakakis2011experience}. In the vein of Experience Driven PCG, this paper presents a method of level generation which attempts to create the same ``game mechanic experience'' as a previous playthrough, while generating levels that are structurally different.

Evolving entire levels to be individually personalized is not trivial. In this paper we propose a method of multi-step evolution. Mini-level ``scenes'' that showcase varying subsets of game mechanics were evolved in previous research~\cite{khalifa2019intentional} using a Constrained Map-Elites algorithm~\cite{khalifa2018talakat} and stored within a publicly available library
~\footnote{https://github.com/LuvneeshM/MarioExperiments/tree/master/scenelibrary}
Our system uses these scenes with an FI-2Pop optimizer to evolve the best ``sequence of scenes,'' with the target being to exactly match a specific player's mechanical playtrace. 

By breaking this problem down into smaller parts, our system is to focus more on the bigger picture of full level creation. Khalifa et al.~\cite{khalifa2019intentional} developed the scene corpus with simplicity and minimalism in mind: a scene's fitness is a direct function of its tile-entropy. Therefore, the levels that our system generates reflect a type of minimalist design, finding the simplest levels which also match the target mechanic sequence.
Generating levels using this method would allow a game system to provide a player with unlimited training for specific mechanic sequences that they struggle on, instead of replaying the same levels over and over again. 
Furthermore, it would allow players to play a diverse set of levels that are still mechanically analogous to the original level that they initially enjoyed playing. 

This paper introduces a level generation method for Mario levels by evolving a sequence of pre-generated, mechanic-focused scenes. Section \ref{sec:background} details the Mario AI Framework within which this research was done as well as an overview of PCG research for games. In Section \ref{sec:methods}, we describe the new generation method and how we evaluate it. Section \ref{sec:results} contains the results of our experiments and examples of the generated levels themselves, and  Section \ref{sec:discussion} discusses the success of the work and how this method could be used in other systems.

\section{Background}\label{sec:background}
\subsection{Search-Based Procedural Content Generation}
Search-based PCG is a generative technique that uses search methods to find good content~\cite{togelius2011search}. In practice, evolutionary algorithms are often used, as they can be applied to many domains, including content and level generation in video games. Search-based PCG has been applied in many different game frameworks, including the General Video Game AI framework~\cite{perez2016general} and PuzzleScript~\cite{khalifa2015automatic}. Ashlock et al.~\cite{ashlock2010automatic,ashlock2011search} explored different evolutionary techniques for puzzle generation of various difficulties and stylized cellular automata evolution for cave generation~\cite{ashlock2015evolvable}. 
Shaker et al.~\cite{shaker2013evolving} generated levels for the puzzle game \emph{Cut the Rope} (ZeptoLab, 2010) by optimizing for playable levels. The research in this paper drew inspiration from several previous works~\cite{mcguinness2011decomposing,dahlskog2014multi,togelius2013patterns} where the authors developed evolutionary processes which divided level generation into separate micro- and macro-evolutionary problems in order to obtain better results for large game levels. 

\subsection{Mario AI Framework}
\emph{Infinite Mario Bros.}~\cite{persson2008infinite} is a public domain clone of Super Mario Bros (Nintendo 1985). Just like in the original game, the player controls Mario by moving horizontally on two-dimensional maps towards a flag on the right hand side. Players can walk, run, and jump to traverse the level. Depending on Mario's mushroom consumption, the player may withstand up two 2 collisions with an enemy before losing and even shoot fireballs.
Players may also lose if Mario falls down a gap. Infinite Mario Bros doesn't use a set levels like in the original Super Mario Bros instead it generates infinite levels using certain rules. For example: gaps can't be wider than 5 tiles because it will be too hard to jump.

The Mario AI framework~\cite{karakovskiy2012mario} is an artificial intelligence research environment built form Infinite Mario Bros. The framework supports several AI players, level generators, and huge set of levels (including the original mario levels). This framework has been used for AI competitions in the past~\cite{karakovskiy2012mario,togelius2013mario}, as well as research into AI gameplay~\cite{karakovskiy2012mario} and level generation~\cite{shaker20112010,sorenson2010towards}.

\subsection{Level Generation in Mario AI}
Several level generators exist in the Mario AI Framework as either artifacts of past competitions or from research projects done independently of them. Competitions for level generation were hosted in 2010 and 2012~\cite{togelius2013mario}. A detailed list of these generators was created by Horne et al.~\cite{horn2014comparative}, which we describe below in addition to other generators written outside the competition. 

The \textit{Notch} and the \textit{Parameterized-Notch} generators add game elements using probability and performs playability checks on the generated levels~\cite{shaker2011feature}. The \emph{Hopper} generator adds to this by adapting content to previous player performance, resulting in dynamic levels that become easier or harder~\cite{shaker20112010}. Using grammars, the \emph{Launchpad} generator generates levels with rhythmic constraints~\cite{smith2011launchpad}. Similar to the generator in this paper, the \textit{Occupancy-Regulated Extension} generator stitches small hand-authored level templates together to create complete levels~\cite{shaker20112010}. A core difference between the Occupancy-Regulated Extension generator and the work in this paper is that the level templates here are generated rather than hand-made.
The \textit{Pattern-based} generator uses single column \emph{slices} taken from the original Super Mario Bros (Nintendo 1985) to evolve levels, optimizing for slice variety~\cite{dahlskog2013patterns}. By expanding design patterns, the \textit{Grammatical Evolution} generator evolves levels by maximizing the number of items in the level and the minimizing conflicts between the placement of these items.

Several research projects have attempted to generate game levels personalized for a specific player or playstyle. The educational game \textit{Refraction} (Center for Game Science at the University of Washington 2010) generates levels using answer set programming to target particular level features~\cite{smith2012case}. 
Khalifa et al.~\cite{khalifa2018talakat} evolved levels for bullet hell games via constrained Map-Elites, a hybrid evolutionary search, using automated playing agents that mimiced different human playstyles. Within the Mario AI Framework, Green et al.~\cite{green2018generating} proposed a method to automatically generate mini-levels in the Mario AI Framework, called ``scenes'', which focused on requiring the player to trigger a specific mechanic in order to win. By evolving scenes using constrained Map-Elites, Khalifa et al.~\cite{khalifa2019intentional} built upon this research and was able to generate a multitude of levels that featured various subsets of game mechanics. The corpus of ``scenes'' that were publicly available from Khalifa et al.'s project is the same library used in this paper (see Section \ref{sec:experiments}). 

\section{Methods}\label{sec:methods}
The method proposed in this paper uses a technique that could most appropriately be called ``level stitching''. A set of pre-evolved miniature levels can be ``stitched'' together to create a longer level and thus a different experience.
These mini-levels, or scenes, were originally evolved to heavily promote the use of sub-sets of game mechanics in a previous project~\cite{khalifa2019intentional}.
Using a target mechanic playtrace as input, our system is able to generate levels that are mechanically analogous to the original.

\begin{table}
    \centering
    \begin{tabular}{|p{.2\linewidth}|p{.7\linewidth}|}
        \hline
        Name & Description \\
        \hline
        Jump & tracks if Mario left the ground\\
        Low Jump & tracks if Mario made a low jump (have a small y axis change less than a certain threshold)\\
        High Jump & tracks if Mario made a high jump (have a high y axis change greater than a certain threshold)\\
        Short Jump & tracks if Mario made a short jump (have a small x axis change less than a certain threshold)\\
        Long Jump & tracks if Mario made a long jump (have a large x axis change greater than a certain threshold)\\
        Stomp Kill & if Mario kills an enemy by jumping on top of it\\
        Shell Kill & if Mario kills an enemy by pushing a Koopa shell\\
        Fall Kill & if a enemy dies because it fell off the game screen\\
        Mode & if Mario changed his mode (small, big, and fire)\\
        Coin & if Mario collected a coin\\
        Brick Block & if Mario bumped into a brick block\\
        Question Block & if Mario bumped into a question mark block\\
        \hline
    \end{tabular}
    \caption{A list of the game mechanics that may be used during a play.}
    \label{tab:usedMechanics}
\end{table}

A mechanic is a game event triggered by player input, or perhaps from another game object, which changes the game state~\cite{sicart2008defining}. For example: jumping over a gap and getting hit by a bolder are both game mechanics. Table~\ref{tab:usedMechanics} shows all the game mechanics in the Mario AI Framework that our system can track during a playthrough, AI or human. It is possible for more than one mechanic to occur at the same time, such as jumping and bumping a brick block. Note that many of these mechanics relate to Mario jumping. Each jump correlates to a different player input and therefore different trajectories for Mario relative to his starting point of the jump. Thus, it is necessary to track many different kinds of jumps, as jumping makes up a core part of Mario gameplay (Table \ref{tab:inputInfo}).

\begin{figure}
    \centering
    \includegraphics[width=0.6\linewidth]{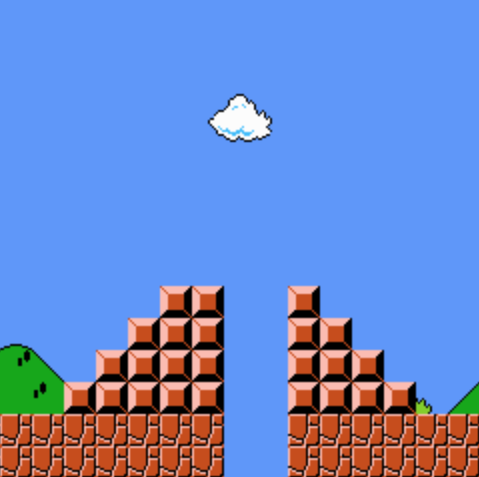}
    \caption{A Super Mario Bros scene where Mario needs to jump from the first pyramid to the second to overcome it.}
    \label{fig:smb_scene}
\end{figure}

A scene is a small area of the level that encapsulates a certain idea~\cite{anthropy2014game}. Figure~\ref{fig:smb_scene} shows a scene from level 1-1 in Super Mario Bros (Nintendo, 1985) where Mario needs to jump from the first pyramid to the second to overcome that scene. For the purposes of this project, a ``scene'' can represent a subset of game mechanics showcased within during gameplay. Figure \ref{fig:smb_scene}, for example, would showcase one or more of the jump mechanics. A scene library should encapsulate all different types of mechanic combinations, from completely flat scenes with no showcased mechanics to scenes where every single game mechanic shown in Table \ref{tab:usedMechanics} is showcased.

This system evolves a full Mario level using the Feasible Infeasible 2-Population (FI-2Pop) genetic algorithm~\cite{Kimbrough2008Introducing}. FI-2Pop uses two populations during evolution, where one tries to satisfy certain constraints and the other optimizes using a fitness value. Generated levels must not only be playable (constraint) but also be optimized to contain the same mechanic experience as the input playthrough (fitness).  Chromosomes may move freely from the infeasible population to the feasible population if they satisfy the constraints and may move back if they fail. Our system uses elitism between generations, where the top performing feasible chromosomes automatically move on to the next generation.

In the following subsections, we will explain each part of the FI-2Pop (representation, operators, and fitness) and how our system uses it to stitch pre-generated scenes into playable Mario levels that contain a desired game mechanic sequence.

\subsection{Chromosome Representation}
A level consists of a varying length of ``scenes'' stitched together. An FI-2Pop chromosome is therefore synonymous with its level representation. A single scene within this chromosome is not limited to only having one mechanic labeled within it. In fact it may be labeled with multiple mechanics. Having scenes that contain multiple mechanics will allow the generator to generate levels that are more condensed.

\subsection{Genetic Operators}
We use two genetic operators: Mutation and Crossover. For crossover, we use a two point crossover with variable length. We pick two points in each parent and swap the center area between these two points. This crossover allows to increase the length of the level or decrease it or swap any number of scenes from 1 scene to the whole level. On the other hand, we use five different types of mutations:
\begin{itemize}
    \item \textbf{Delete:} delete a scene.
    \item \textbf{Add:} add a random scene adjacent to a scene. The random scene is selected with probability inversely proportional to number of mechanics using rank selection.
    \item \textbf{Split:} split a scene in half and replace it with a left and new right scene, randomly selecting half the mechanics to go in a new left scene and the rest to go in the right.
    \item \textbf{Merge:} add the mechanics of a scene to the left or right scene, then replace both scenes with one from the corpus that has the combined list of mechanics.
    \item \textbf{Change:} changes a scene with another random scene. The random scene is selected with probability inversely proportional to number of mechanics using rank selection.
\end{itemize}
To apply the mutation, first the system has to select a scene to apply one of the previously discussed operators. The scene is selected such that scenes with more mechanics have a higher chance to be selected than scenes with low mechanics.

\subsection{Constraints and Fitness Calculation}
FI-2Pop uses a dual-population system, the infeasible population and feasible population. The infeasible population tries to make the chromosomes satisfy constraints (like making sure the level is playable), while the feasible population tries to make sure that the mechanics in the new levels are similar to the input mechanic sequence.

To calculate the constraints, we run an artificial agent on each chromosome for $N$ times where $N$ is a preset constant. Based on the result we calculate the constraint value using the following equation:
\begin{equation}
    C = \begin{cases}
        \frac{1}{N}\sum_{i=1}^{N}\frac{d_{i}}{d_{level}} &\text{if $\frac{1}{N}\sum_{i=1}^{N} w_{i} < p$} \\
        1 &\text{if $\frac{1}{N}\sum_{i=1}^{N} w_{i} \geq p$}
    \end{cases}
\end{equation}
where $d_{i}$ is the distance traveled by the A* agent on the level on the $i^{th}$ iteration, $d_{level}$ is the maximum length of the level, $w_{i}$ is equal 1 if the agent reached the end of the level on the $i^{th}$ run, and $p$ is the threshold percentage.

\begin{figure*}[t]
    \centering
    \includegraphics[width=.9\textwidth]{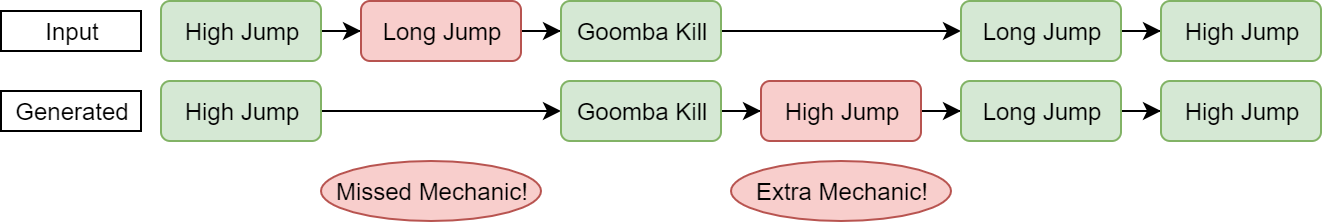}
    \caption{An example of missing and extra mechanic faults. The Long jump in the input playthrough is missing in the newly generated map's playthrough. The generated playthrough also contains an extra high jump not included in the input.}
    \label{fig:fault_example}
\end{figure*}

To calculate the fitness, the system finds the single agent playthrough in which the agent not only won the level but also triggered the least fired mechanics. The level is assigned an initial score $S$, which is then decremented based on the number of ``faults''.
A fault is a mechanic sequence mismatch between the input sequence and the newly generated agent playthrough, either as an extra mechanic placed between a correct subsequence of mechanics, or else as a missing mechanic that would create an otherwise correct subsequence. Figure \ref{fig:fault_example} displays an example of a missing mechanic and an extra mechanic between the input and generated agent playthroughs. Our system uses a sequence matching algorithm to calculate fault counts, as shown by Algorithm \ref{alg:fault_algorithm}.  

In Algorithm \ref{alg:fault_algorithm} we loop over the target sequence and find the first occurrence of each target mechanic in a sub-array of the generated mechanics list, initially from the beginning to the end of the array. If the mechanic is not found, we increment a counter tracking the number of missed mechanics. Otherwise, the mechanic is found and the index of the mechanic is the number of extra mechanics between two matching mechanics. The pointer of the generated mechanic sequence is moved to point at this new position to continue the loop.

\begin{algorithm}
\SetAlgoLined
\textbf{GIVEN: generatedSeq, targetSeq}\\
pTarget = $0$; pGenerated = $0$\;
extraMechs = $0$; missedMechs = $0$\;
 \For{pTarget < len(targetSeq), pTarget += 1} {
    targetMechanic = targetSeq[pTarget]\;
    genSubList = generatedSeq.subList(pGenerated,len(generatedSeq))\;
    mechanicIndex = genSubList.indexOf(targetMechanic)\;
    \eIf{mechanicIndex == -1}{
        missedMechs += 1\;
    }{
        pTarget += mechanicIndex + 1\;
	    extraMechs += mechanicIndex\;
    }
    \If{pGenerated >= len(generatedSeq)}{
    	pTarget += 1\;
		break\;
    }
 }
 \Return{extraMechs, missedMechs}
 \caption{Calculating fault count in a chromosome}
 \label{alg:fault_algorithm}
\end{algorithm}

As we evaluate a level's fitness based on what it misses and for having excess, it is possible for a level to have a negative fitness score. The fitness function is calculated based on the following equation:
\begin{equation}
\label{eqn:fitnessScore}
\begin{split}
P_{missed} = W_{missed} * M_{missed}\\
P_{extra} = a * \tanh(b * M_{extra}) + c\\
F = S - (P_{missed} + P_{extra} * (S - P_{missed}))
\end{split}
\end{equation}
where $P_{missed}$ is the penalty of missed mechanics, $P_{extra}$ is the penalty of extra mechanics, $M_{missed}$ is the number of missed mechanics, $M_{extra}$ is the number of extra mechanics, $S$ is the initial starting value, and $a$, $b$, $c$, and $W_{missed}$ are predefined weights.

\section{Experiments}\label{sec:experiments}
To test our algorithms, we generate levels using three original Super Mario levels as targets (World 1-1, World 4-2, and World 6-1, shown as ``original'' in Figures \ref{fig:levels_original_1-1}, \ref{fig:levels_original_4-2}, and \ref{fig:levels_original_6-1}). The Robin Baumgarten A* algorithm, which was developed for the first Mario AI competition~\cite{Togelius2009The}, is run once on each of the three levels. The resulting mechanic sequences are collected and used as targets. 

The evolutionary algorithm uses a population of $250$ chromosomes each generation, with a $70\%$ rate for crossover, a $20\%$ rate for mutation, and $1$ elite. A single chromosome is initialized using a random scene picker, which selects anywhere between $5$ to $25$ scenes with which to populate the level. The scene corpus is taken from the results of \textit{Intentional Computational Level Design} \cite{khalifa2019intentional}. The corpus~\footnote{hidden for peer review} contains a total of $1691$ Mario scenes, containing nearly every possible combination of Mario mechanics, with an average of given combination. Each scene for the single level for the chromosome is randomly selected from the corpus based on the assigned number of mechanics to it. The number of mechanics for each scene is sampled from a Gaussian distribution with mean equal to the average number of mechanics in the target and standard deviation of 1.


To compare our techniques, we compare the results from our algorithms with two baselines. The random baseline generates levels with a random scene count between 5 and 25 where each scene is picked randomly from the corpus, independent of any available information. The greedy baseline generates levels with a random scene count between 5 and 25 similar to that from the random baseline. The scenes used to generate the level are then selected such that the resulting level maximizes the number of matched mechanics based on each scenes labeled mechanics in the corpus. For both baseline generators, we generate levels with their respective methods until a total of 20 playable levels for each world is generated. A playable level is defined as a level in which the A* Agent was able to complete during the time of level creation. Table \ref{tab:inputInfo} displays information about the mechanic makeup of the input playtraces, broken down by level.

\begin{table}[]
\begin{tabular}{|l|l|l|l|l|}
\hline
\textbf{Level} & \textbf{Total} & \textbf{Unique} & \textbf{Most Freq} & \textbf{Least Freq} \\ \hline
1-1            & 57             & 8               & Jump (18)          & Stomp Kill (1)      \\ \hline
4-2            & 96             & 8               & Jump (29)          & Stomp Kill (2)      \\ \hline
6-1            & 66             & 6               & Jump (22)          & Coin (1)    \\ \hline
\end{tabular}
\caption{Metrics about the input target playtraces, including which mechanics were most and least frequently triggered. Note that ``Jump'' makes up a large amount of triggered mechanics.}
\label{tab:inputInfo}
\end{table}

\section{Results}\label{sec:results}

\begin{figure*}
    \centering
    \begin{subfigure}[t]{\linewidth}
        \centering
        \includegraphics[width=\textwidth]{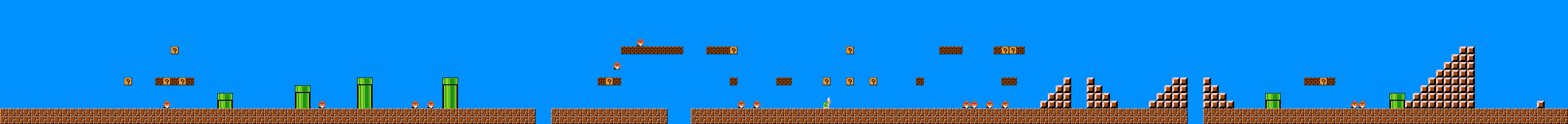}
        \caption{Level 1-1 original}
        \label{fig:levels_original_1-1}
    \end{subfigure}
    \begin{subfigure}[t]{\linewidth}
        \centering
        \includegraphics[width=\textwidth]{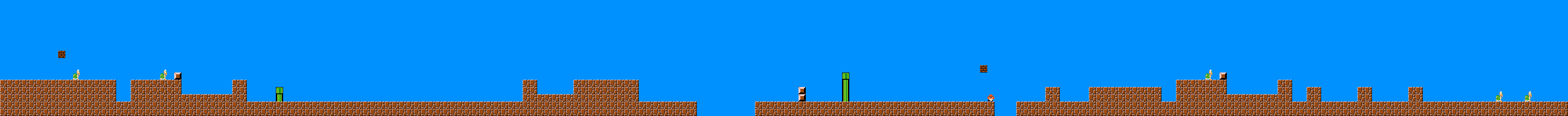}
        \caption{Level 1-1 greedy}
        \label{fig:levels_greedy_1-1}
    \end{subfigure}
    \begin{subfigure}[t]{\linewidth}
        \centering
        \includegraphics[width=\textwidth]{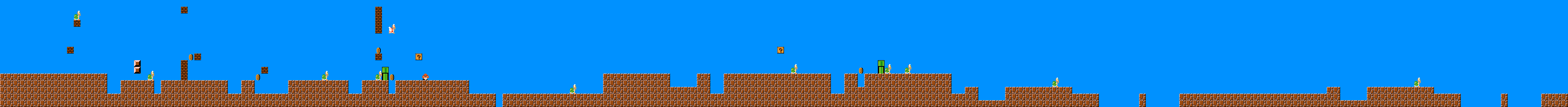}
        \caption{Level 1-1 evolved}
        \label{fig:levels_evo_1-1}
    \end{subfigure}
    \begin{subfigure}[t]{\linewidth}
        \centering
        \includegraphics[width=\textwidth]{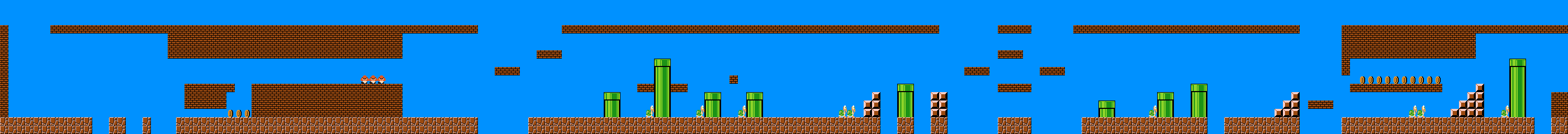}
        \caption{Level 4-2 original}
        \label{fig:levels_original_4-2}
    \end{subfigure}    
    \begin{subfigure}[t]{\linewidth}
        \centering
        \includegraphics[width=\textwidth]{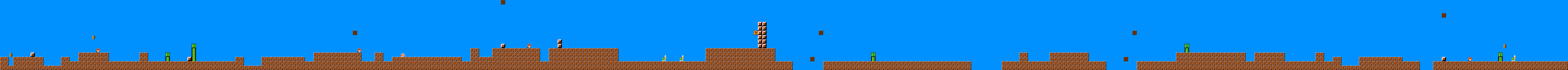}
        \caption{Level 4-2 greedy}
        \label{fig:levels_greedy_4-2}
    \end{subfigure}
    \begin{subfigure}[t]{\linewidth}
        \centering
        \includegraphics[width=\textwidth]{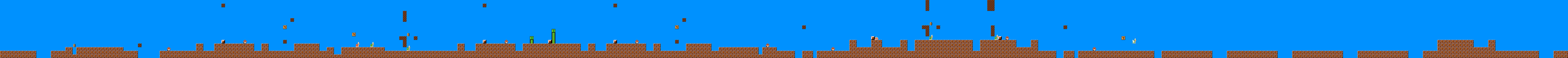}
        \caption{Level 4-2 evolved}
        \label{fig:levels_evo_4-2}
    \end{subfigure}
    \begin{subfigure}[t]{\linewidth}
        \centering
        \includegraphics[width=\textwidth]{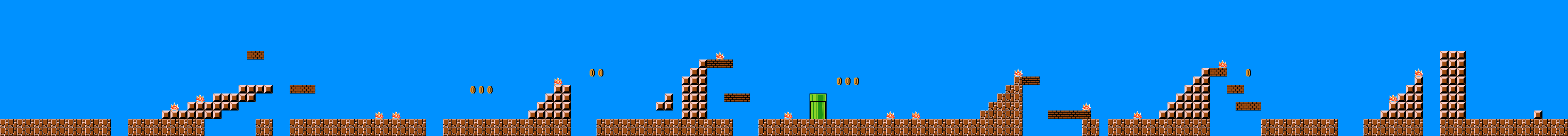}
        \caption{Level 6-1 original}
        \label{fig:levels_original_6-1}
    \end{subfigure}
    \begin{subfigure}[t]{\linewidth}
        \centering
        \includegraphics[width=\textwidth]{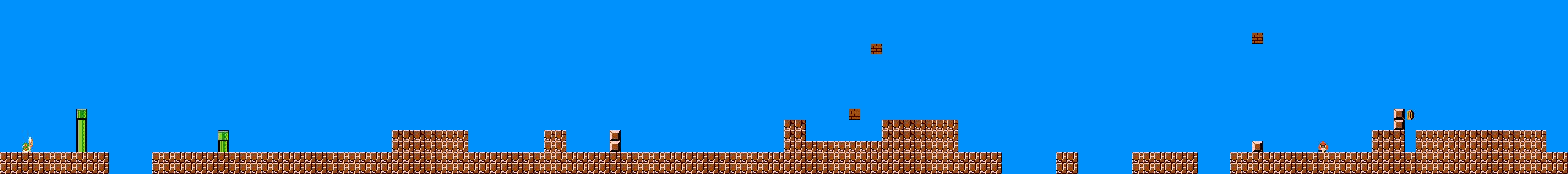}
        \caption{Level 6-1 greedy}
        \label{fig:levels_greedy_6-1}
    \end{subfigure}
    \begin{subfigure}[t]{\linewidth}
        \centering
        \includegraphics[width=\textwidth]{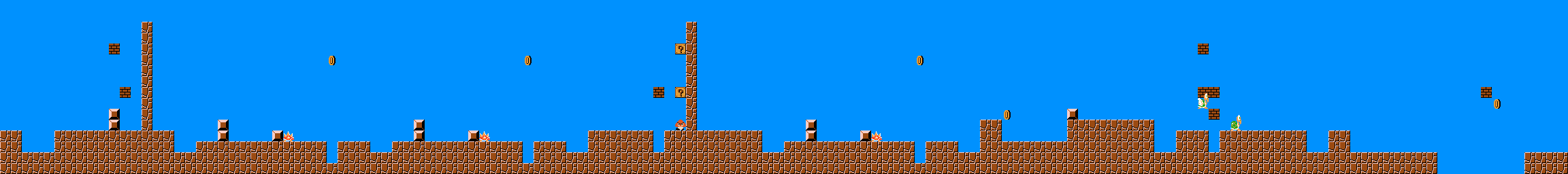}
        \caption{Level 6-1 evolved}
        \label{fig:levels_evo_6-1}
    \end{subfigure}
    \caption{A random sampling of greedy-generated and system-evolved levels, compared to their original equivalents}
    \label{fig:playthroughResults}
\end{figure*}

Figure~\ref{fig:playthroughResults} shows the each of the original levels from Super Mario Bros (Nintendo, 1985) and its greedy and evolved counterparts. Both greedy and evolved levels have less graphical variance, a result of the lack of diversity in the scenes they stitched together. For example, a scene containing a pipe 3 tiles tall is same as having 3 breakable brick tiles stacked on top of each other. 
The entropy function used to evolve these scenes~\cite{khalifa2019intentional} most likely negatively impacts level diversity, as it aims to simply levels and create uniform spaces. Most of the generated levels (greedy or evolution) seem flatter than their original counterparts. The entropy function used for evolving the scenes~\cite{khalifa2019intentional} implies a pressure for lower height variance, and therefore impacts levels created with the scenes in a similar way. In order to enforce a specific mechanic sequence, the evolved levels seem to have taken this to the extreme, by using scenes with tall vertical walls only 1-tile wide to create opportunities for multiple jumps in a row.

\subsection{Evolution Performance}
\begin{figure}
    \centering
    \includegraphics[width=.9\columnwidth]{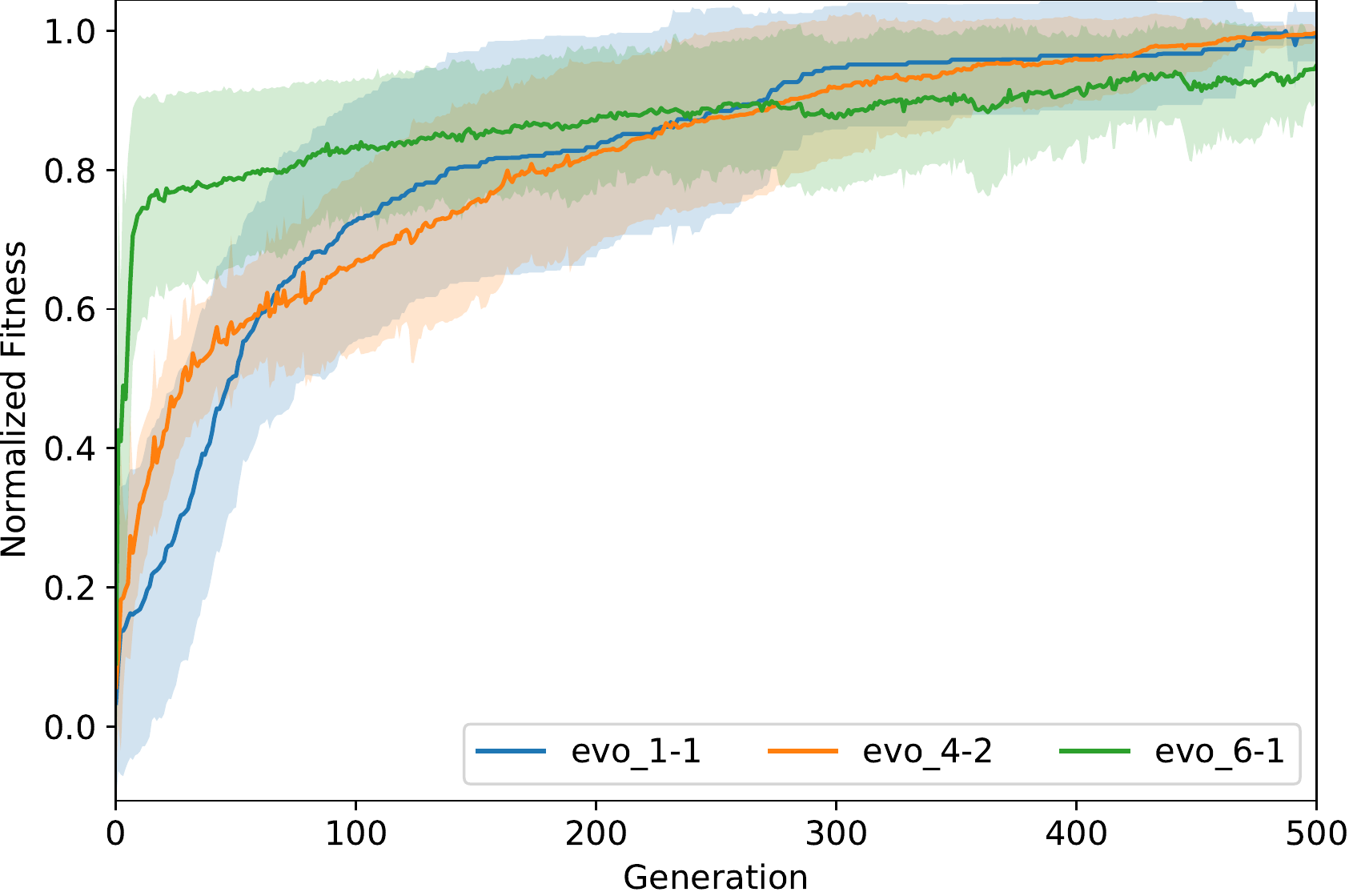}
    \caption{Tracking evolution performance throughout generations across all three target levels for all generators.}
    \label{fig:evolutionMetrics}
\end{figure}

Figure \ref{fig:evolutionMetrics} displays the normalized fitness curves of the evolutionary generators, averaged over the 20 generated levels per target playtrace. The shaded curves display the standard deviation of each curve. The fitness curve of the level 1-1 experiment suggests that the mechanic sequence was easy to match, as we see near linear performance increases until generation $80$, after which fitness improvement slows down. Similarly, 6-1 evolution climbs rapidly within the first 20 generations on average, however with a higher deviation. 4-2 slows to sub-linear gains after a period of rapid growth ending around generation $40$, with a rather small deviation relative to the others.

\subsection{Level Playability}

\begin{table}
    \centering
    \begin{tabular}{|l|r|}
        \hline
        Experiment &  Playability\\
        \hline
        Original Levels & 52\% \\
        Random Levels & 10.75\% \\
        Greedy World 1-1 & 28.5\% \\
        Greedy World 4-2 & 26.25\% \\
        Greedy World 6-1 & 25\% \\
        Evolution World 1-1 & 100\% \\
        Evolution World 4-2 & 99.5\% \\
        Evolution World 6-1 & 87.25\% \\
        \hline
    \end{tabular}
    \caption{The percentage of times the Robin Baumgurten A* algorithm wins this set of levels over 20 runs for each level.}
    \label{tab:playability}
\end{table}

To calculate playability for each group of levels, we run Robin Baumgarten's A* agent~\cite{togelius20102009} $20$ times per level and average the results over the whole group of levels.
Table~\ref{tab:playability} shows the playability percentages of each of these groups. We find it notable that the A* can only win 52\% of the original levels, which includes all the levels from the original Super Mario Bros except for the underwater levels and castle levels. This demonstrates that the A* is not a perfect algorithm and not able to beat every level every time. 
The evolved 1-1, 4-2, and 6-1 levels all have close to 100\% playability. In contrast, greedy stitching seems to make poor quality levels in terms of playability ($25-28.5\%$). Random stitching predictably creates barely playable levels ($10\%$).

\subsection{Mechanic Similarity}
\begin{figure*}
    \centering    \includegraphics[width=.9\textwidth]{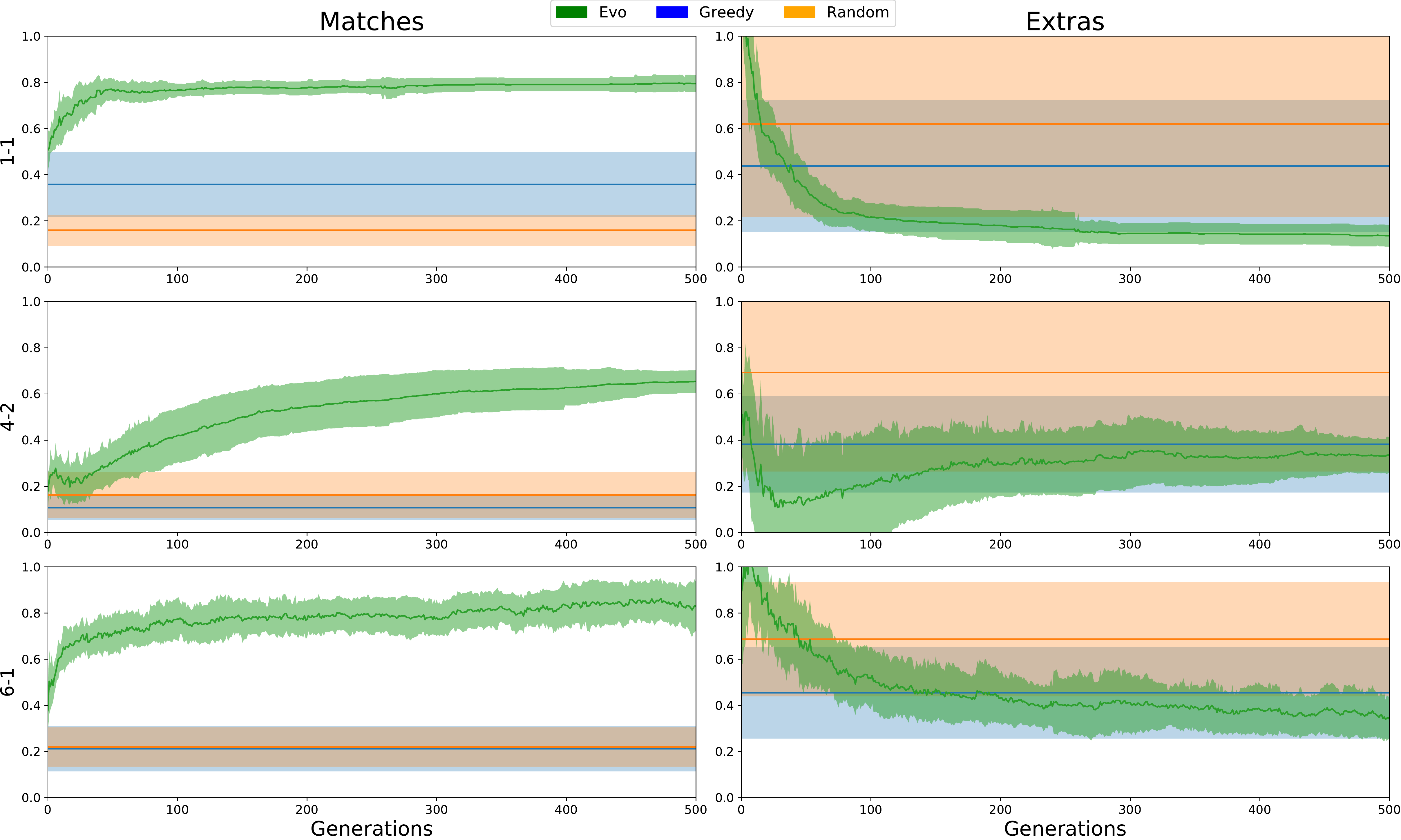}
    \caption{Tracking mechanic statistics throughout generations across all three target levels for all generators.}
    \label{fig:mechanicMetrics}
\end{figure*}

Figure \ref{fig:mechanicMetrics} displays the mechanic evaluation across all three target levels for all generators. ``Matches'' and ``Extras'' are normalized using the \emph{total} value from Table \ref{tab:usedMechanics} for each level. As fitness is impacted by matching mechanics it makes sense that the evolutionary generated levels go up over time, just as fitness does. Across all three levels, the evolution agent far outperforms both the greedy and random generators. This is also true for the extra mechanics (which are minimized) on 1-1 and 6-1. However, on 4-2 the greedy generator seems to add more extra mechanics after a brief drop ending around generation $50$ which might be inevitable to increase the matched mechanics. This stabilizes around generation $200$, which is also when the match mechanic count stabilizes.

\subsection{Structural Diversity}\label{sec:experiments_diversity}
To measure the diversity between the generated content, we used Tile Pattern KL-Divergence (TPKLDiv) calculations~\cite{lucas2019tile} to measure the structural similarity between the 20 generated levels. For each possible pair of levels, we measure the TPKLDiv for each pair of levels from which we pick the minimum TPKLDiv value for each level (the worst case). We end up with 20 values, we take their average to formulate that technique-target combination's TPKLDiv score. There are only 15 original Mario levels included in the framework, so we use those 15 TPKLDiv values to find that group's average score.

\begin{table}
    \centering
    \begin{tabular}{|l|r|}
        \hline
        Experiment & TPKLDiv\\
        \hline
        Original Levels & $0.715 \pm 0.410$ \\
        Random Levels & $0.697 \pm 0.265$ \\
        Greedy World 1-1 & $0.675 \pm 0.228$ \\
        Greedy World 4-2 & $0.648 \pm 0.181$ \\
        Greedy World 6-1 & $0.648 \pm 0.172$ \\
        Evolution World 1-1 & $0.269 \pm 0.127$ \\
        Evolution World 4-2 & $0.264 \pm 0.094$ \\
        Evolution World 6-1 & $0.348 \pm 0.117$ \\
        \hline
    \end{tabular}
    \caption{Tile Pattern KL-Divergence calculation between the generated levels for each type, plus standard deviation. We included a comparison for all the original levels as well as a reference point.}
    \label{tab:diveristy_within}
\end{table}

We use a 3x3 tile pattern window to calculate the TPKLDiv. Table~\ref{tab:diveristy_within} shows the TPKLDiv values between the 20 generated levels. It is obvious that every technique has a lower diversity score than the original levels of Super Mario Bros since the random levels have lower TPKLDiv value which show that the actual scenes have fewer diversity compared to the original Mario levels. Also, the levels generated for World 1-1 have low diversity relative to the others. In the context of it being the very first level of the game with the most basic and simple mechanics for new players, this diversity score makes sense. This makes it harder to find more diverse structural levels. It is surprising that World 4-2 has less diversity than World 1-1. The World 4-2 playtrace has so many fired mechanics, it might be more difficult to find a playable level with that amount of mechanics in 25 scenes especially when scenes with small amount of mechanics are being selected more often. One last note, the greedy algorithm having higher diversity than all the evolution levels but at the same time they have less playable levels compared to the evolved levels.

\begin{table}
    \centering
    \begin{tabular}{|l|r|}
        \hline
        Experiment & TPKLDiv\\
        \hline
        \hline
        Random Levels 1-1 & $2.941 \pm 1.005$ \\
        Greedy World 1-1 & $2.636 \pm 0.795$ \\
        Evolution World 1-1 & $1.601 \pm 0.573$ \\
        \hline
        \hline
        Random Levels 4-2 & $2.942 \pm 1.005$ \\
        Greedy World 4-2 & $3.329 \pm 0.647$ \\
        Evolution World 4-2 & $1.997 \pm 0.466$ \\
        \hline
        \hline
        Random Levels 6-1 & $2.942 \pm 1.005$ \\
        Greedy World 6-1 & $2.601 \pm 0.577$ \\
        Evolution World 6-1 & $1.505 \pm 0.404$ \\
        \hline
    \end{tabular}
    \caption{The average Tile Pattern KL-Divergence calculation between the generated levels and their corresponding level, plus standard deviation. We compared the random levels to each of the target levels as a reference point.}
    \label{tab:diveristy}
\end{table}

Similar to the previous experiment, we compared each of the generated 20 levels (including random and greedy) to its corresponding original level using TPKLDiv calculation with a 3x3 window. This is done to create a metric measuring how different the generated levels are from their original counterparts. Table~\ref{tab:diveristy} shows the results from these calculations. The random generator's levels have the largest TPKLDiv except for in World 4-2 where we were surprised to find that greedy generator has a higher value. This might be due to the greedy generator creating longer levels in reaction to the longer length of the input mechanic sequence for that level. The evolved levels have nearly half the TPKLDiv value of the random generator levels except for World 4-2, also probably in response to that playtrace's mechanic count. 
To reference the findings presented in Table \ref{tab:diveristy_within}, 4-2 levels seem to contain only small amounts of diversity between themselves within their generation group but show an increased difference between themselves and their original counterpart.

\section{Discussion}\label{sec:discussion}

\begin{figure}
    \centering
    \includegraphics[width=.9\columnwidth]{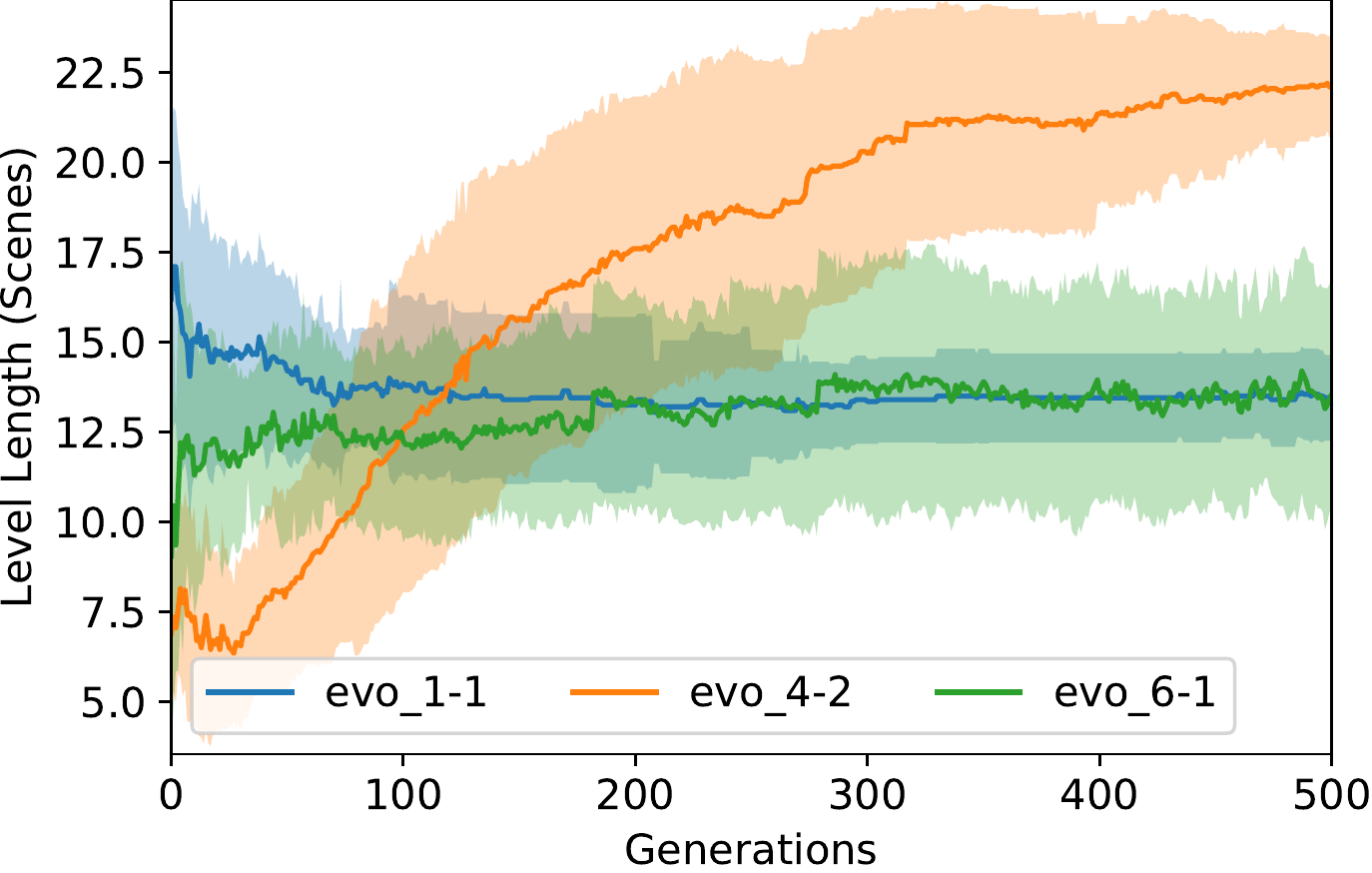}
    \caption{Progression of averaged level lengths for all three observed levels throughout all generations for the evolutionary generator}
    \label{fig:scene_lengths}
\end{figure}

Looking to results as show by Figure~\ref{fig:mechanicMetrics}, it is clear the FI-2Pop generator outperformed the baselines in the defined terms of matching the mechanic sequence of the input.
As we allow for the length of the levels to vary within a defined range, it is important to observe the convergence of said level lengths. A typical level from Super Mario is 14 scenes in length and the levels generated for level 1-1 and level 6-1 converge toward and hover around that length. However, the generator for level 4-2 converges to a length of roughly 23 scenes, nearly $1.64$ times that of the other 2 observed levels. We hypothesis the reason behind this influx in scene length is due to sheer number of mechanics present in the original level 4-2. Figure \ref{fig:scene_lengths} shows level 4-2 to have the most number of mechanics triggered, close to $1.7$ times the amount of level 1-1 and to $1.5$ times the amount for level 6-1. We believe, the generator could not guarantee levels where the input mechanic sequence to occur in the given length and thus favored levels with more scenes. By spreading out the mechanics, it allows the generator to better guarantee generating levels with a higher likelihood of being aligned to the input from a mechanics sequence standpoint. This is evident as the number of matches increases as the overall length of the level increases for level 4-2. Since the overall length of level 4-2 increases, the likelihood of additional mechanics occurring between the wanted mechanics also increases, explaining the rise in extra mechanics for the generated levels for 4-2.

\begin{figure}
    \centering
    \includegraphics[width=.7\linewidth]{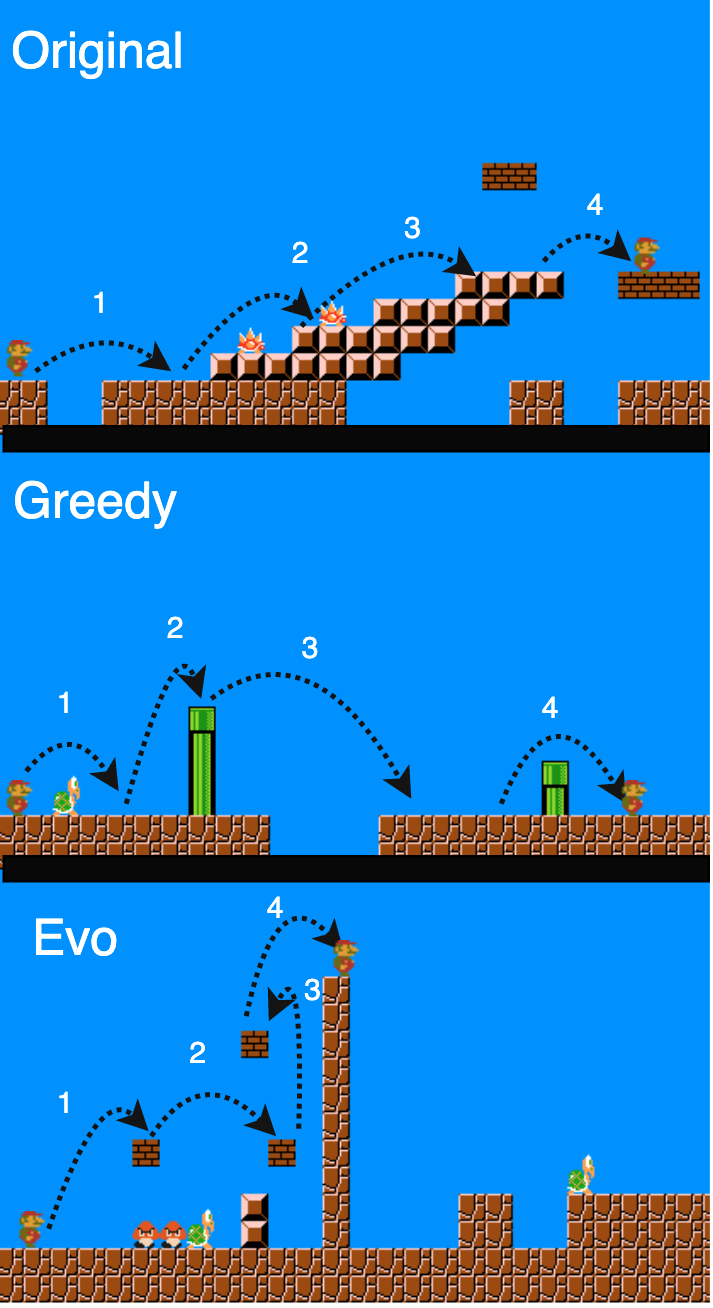}
    \caption{An example of a 4 jump sequence in the original 6-1 level, and how the two generators try to copy it.}
    \label{fig:jump_example}
\end{figure}

We observe that our evolutionary generator is influenced to ensure mechanics from the input sequence are forced to happen in their particular order in the generated levels. That is to say that evolution is driven by the matching pressure in the fitness function to guarantee the agent had no other choice but to perform certain mechanics before progressing forward. Figure \ref{fig:jump_example} shows an example of this with a zoom into a subsection of 6-1 from the original level, greedy generated level and evolved level. The original level requires the agent to perform a 4-jump sequence in order to progress forward in the level. A similar manner of triggered mechanics can be seen in both the greedy and evolved levels. The greedy generator simply places scenes next to each other in which jumps occur. However, it cannot guarantee or force the agent to perform the jumps outside of having a strong likelihood of the jumps occurring. For example, the if the agent is Big Mario in the greedy example, they could collide with the Koopa before jumping onto the pipe instead of jumping over the Koopa. Looking to the evolved example, we see there is a long wall that acts as a hard gate, blocking the agent from progressing forward. The only way the agent can continue forward in the level is by first performing the 3 jumps to achieve the required height to then jump over the long wall. The generator forces the agent into performing the 4-jump sequence to mimic the mechanic sequence from the input. 

The corpus from Khalifa et al.~\cite{khalifa2019intentional} was built using a fitness function that minimized entropy. This minimalist pressure is reflected on the levels built using this corpus. In Figure \ref{fig:jump_example}, we can see how the 2-gap-4-step level structure with spikys that requires 4 jumps to overcome can be boiled down to just a Koopa turtle, a pipe, a gap, and a little wall in the greedy level. The evolutionary generator, driven by matching pressure, shrinks this down into a large column and a few floating blocks to all but guarantee the agent has no other choice but this mechanic sequence. In a way, the evolution method is performing a type of minimalist level generation, creating the simplest levels which are mechanically analogous to the original.

Based on the results of Section \ref{sec:experiments_diversity} and Table \ref{tab:diveristy}, we hypothesize that a mechanical sequence populated with small amounts of relatively simple mechanics (Evolution World 1-1) has only a small range of mechanically similar cousins. It seems that having a large population of mechanics also makes it difficult to curate diversity (World 4-2). Its possible that World 6-2 represents a sweet spot in terms of diversity. In future work, we'd like to test this theory using a more diverse array of scenes and levels.

\section{Conclusion}\label{sec:conclusion}
In this paper, we explore a means to automatically generate personalized content by stitching pre-generated scenes to construct levels for Super Mario. We compare the sequence in which mechanics occur in the playthrough of the Robin Baumgarten A* agent on three unique levels from the original Super Mario (1-1, 4-2 and 6-1) to the sequences the same A* agent would take in the generated levels from the various methods to judge the success of the experiments. We use the FI-2Pop evolution algorithm, with the focus of developing winnable levels that are mechanically analogous to the original level, and we find that it is able to match the sequence much better than either baseline. Although both baselines and the new method have lower diversity among themselves than the original Mario levels, this is most likely a result of the entropy pressure during the scene generation process.

While these results show to be promising, this work can be further expanded to examine playthroughs from personified agents or human testers to generate levels unique for them. We would like to observe how the generated levels for various groups of users differ in what the focus of the level becomes based off how a user plays the initial Super Mario level. For example, a generated level could focus on trying get coins in hard to reach places, like areas in which the player needs to reach both a certain height and perform a long jump. With a user study, it would be possible to gain understanding of player preferences between the original level and personalized generated levels.

The corpus~\cite{khalifa2019intentional} used in this work was built to minimize entropy. As we discussed, this causes levels built with it to exhibit minimalist design. We question the value of always minimizing and propose a new corpus to be generated for a ``target entropy.'' The original Mario levels were never built to be minimalist, so why should the scenes that the generator uses? A future project could re-generate this corpus using the average entropy found across all original Mario levels.
\begin{acks}
Ahmed Khalifa acknowledges the financial support from NSF grant (Award number 1717324 - ``RI: Small: General Intelligence through Algorithm Invention and Selection.''). Michael Cerny Green acknowledges the financial support of the SOE Fellowship from NYU Tandon School of Engineering.
\end{acks}

\bibliographystyle{ACM-Reference-Format}
\bibliography{sample-base}


\begin{thebibliography}{33}


\ifx \showCODEN    \undefined \def \showCODEN     #1{\unskip}     \fi
\ifx \showDOI      \undefined \def \showDOI       #1{#1}\fi
\ifx \showISBNx    \undefined \def \showISBNx     #1{\unskip}     \fi
\ifx \showISBNxiii \undefined \def \showISBNxiii  #1{\unskip}     \fi
\ifx \showISSN     \undefined \def \showISSN      #1{\unskip}     \fi
\ifx \showLCCN     \undefined \def \showLCCN      #1{\unskip}     \fi
\ifx \shownote     \undefined \def \shownote      #1{#1}          \fi
\ifx \showarticletitle \undefined \def \showarticletitle #1{#1}   \fi
\ifx \showURL      \undefined \def \showURL       {\relax}        \fi
\providecommand\bibfield[2]{#2}
\providecommand\bibinfo[2]{#2}
\providecommand\natexlab[1]{#1}
\providecommand\showeprint[2][]{arXiv:#2}

\bibitem[\protect\citeauthoryear{Anthropy and Clark}{Anthropy and
  Clark}{2014}]%
        {anthropy2014game}
\bibfield{author}{\bibinfo{person}{A. Anthropy} {and} \bibinfo{person}{N.
  Clark}.} \bibinfo{year}{2014}\natexlab{}.
\newblock \bibinfo{booktitle}{\emph{A Game Design Vocabulary: Exploring the
  Foundational Principles Behind Good Game Design}}.
\newblock \bibinfo{publisher}{Pearson Education}.
\newblock
\showISBNx{9780321886927}
\showLCCN{2013956696}


\bibitem[\protect\citeauthoryear{Ashlock}{Ashlock}{2010}]%
        {ashlock2010automatic}
\bibfield{author}{\bibinfo{person}{Daniel Ashlock}.}
  \bibinfo{year}{2010}\natexlab{}.
\newblock \showarticletitle{Automatic generation of game elements via
  evolution}. In \bibinfo{booktitle}{\emph{Computational Intelligence and Games
  Conference}}. IEEE, \bibinfo{pages}{289--296}.
\newblock


\bibitem[\protect\citeauthoryear{Ashlock}{Ashlock}{2015}]%
        {ashlock2015evolvable}
\bibfield{author}{\bibinfo{person}{Daniel Ashlock}.}
  \bibinfo{year}{2015}\natexlab{}.
\newblock \showarticletitle{Evolvable fashion-based cellular automata for
  generating cavern systems}. In \bibinfo{booktitle}{\emph{Computational
  Intelligence and Games Conference}}. IEEE, \bibinfo{pages}{306--313}.
\newblock


\bibitem[\protect\citeauthoryear{Ashlock, Lee, and McGuinness}{Ashlock
  et~al\mbox{.}}{2011}]%
        {ashlock2011search}
\bibfield{author}{\bibinfo{person}{Daniel Ashlock}, \bibinfo{person}{Colin
  Lee}, {and} \bibinfo{person}{Cameron McGuinness}.}
  \bibinfo{year}{2011}\natexlab{}.
\newblock \showarticletitle{Search-based procedural generation of maze-like
  levels}.
\newblock \bibinfo{journal}{\emph{Transactions on Computational Intelligence
  and AI in Games}} \bibinfo{volume}{3}, \bibinfo{number}{3}
  (\bibinfo{year}{2011}), \bibinfo{pages}{260--273}.
\newblock


\bibitem[\protect\citeauthoryear{Dahlskog and Togelius}{Dahlskog and
  Togelius}{2013}]%
        {dahlskog2013patterns}
\bibfield{author}{\bibinfo{person}{Steve Dahlskog} {and}
  \bibinfo{person}{Julian Togelius}.} \bibinfo{year}{2013}\natexlab{}.
\newblock \showarticletitle{Patterns as objectives for level generation}.
\newblock  (\bibinfo{year}{2013}).
\newblock


\bibitem[\protect\citeauthoryear{Dahlskog and Togelius}{Dahlskog and
  Togelius}{2014}]%
        {dahlskog2014multi}
\bibfield{author}{\bibinfo{person}{Steve Dahlskog} {and}
  \bibinfo{person}{Julian Togelius}.} \bibinfo{year}{2014}\natexlab{}.
\newblock \showarticletitle{A multi-level level generator}. In
  \bibinfo{booktitle}{\emph{2014 IEEE Conference on Computational Intelligence
  and Games}}. IEEE, \bibinfo{pages}{1--8}.
\newblock


\bibitem[\protect\citeauthoryear{Green, Khalifa, Barros, Nealen, and
  Togelius}{Green et~al\mbox{.}}{2018}]%
        {green2018generating}
\bibfield{author}{\bibinfo{person}{Michael Green}, \bibinfo{person}{Ahmed
  Khalifa}, \bibinfo{person}{Gabriella Barros}, \bibinfo{person}{Andy Nealen},
  {and} \bibinfo{person}{Julian Togelius}.} \bibinfo{year}{2018}\natexlab{}.
\newblock \showarticletitle{Generating levels that teach mechanics}. In
  \bibinfo{booktitle}{\emph{Procedural Content Generation Workshop}}.
  \bibinfo{publisher}{ACM}.
\newblock


\bibitem[\protect\citeauthoryear{Horn, Dahlskog, Shaker, Smith, and
  Togelius}{Horn et~al\mbox{.}}{2014}]%
        {horn2014comparative}
\bibfield{author}{\bibinfo{person}{Britton Horn}, \bibinfo{person}{Steve
  Dahlskog}, \bibinfo{person}{Noor Shaker}, \bibinfo{person}{Gillian Smith},
  {and} \bibinfo{person}{Julian Togelius}.} \bibinfo{year}{2014}\natexlab{}.
\newblock \showarticletitle{A comparative evaluation of procedural level
  generators in the mario ai framework}. \bibinfo{publisher}{Society for the
  Advancement of the Science of Digital Games}.
\newblock


\bibitem[\protect\citeauthoryear{Karakovskiy and Togelius}{Karakovskiy and
  Togelius}{2012}]%
        {karakovskiy2012mario}
\bibfield{author}{\bibinfo{person}{Sergey Karakovskiy} {and}
  \bibinfo{person}{Julian Togelius}.} \bibinfo{year}{2012}\natexlab{}.
\newblock \showarticletitle{The mario ai benchmark and competitions}.
\newblock \bibinfo{journal}{\emph{Transactions on Computational Intelligence
  and AI in Games}} \bibinfo{volume}{4}, \bibinfo{number}{1}
  (\bibinfo{year}{2012}), \bibinfo{pages}{55--67}.
\newblock


\bibitem[\protect\citeauthoryear{Khalifa, Bontrager, Earle, and
  Togelius}{Khalifa et~al\mbox{.}}{2020}]%
        {khalifa2020pcgrl}
\bibfield{author}{\bibinfo{person}{Ahmed Khalifa}, \bibinfo{person}{Philip
  Bontrager}, \bibinfo{person}{Sam Earle}, {and} \bibinfo{person}{Julian
  Togelius}.} \bibinfo{year}{2020}\natexlab{}.
\newblock \showarticletitle{PCGRL: Procedural Content Generation via
  Reinforcement Learning}.
\newblock \bibinfo{journal}{\emph{arXiv preprint arXiv:2001.09212}}
  (\bibinfo{year}{2020}).
\newblock


\bibitem[\protect\citeauthoryear{Khalifa and Fayek}{Khalifa and Fayek}{2015}]%
        {khalifa2015automatic}
\bibfield{author}{\bibinfo{person}{Ahmed Khalifa} {and} \bibinfo{person}{Magda
  Fayek}.} \bibinfo{year}{2015}\natexlab{}.
\newblock \showarticletitle{Automatic puzzle level generation: A general
  approach using a description language}. In
  \bibinfo{booktitle}{\emph{Computational Creativity and Games Workshop}}.
\newblock


\bibitem[\protect\citeauthoryear{Khalifa, Green, Barros, and Togelius}{Khalifa
  et~al\mbox{.}}{2019}]%
        {khalifa2019intentional}
\bibfield{author}{\bibinfo{person}{Ahmed Khalifa},
  \bibinfo{person}{Michael~Cerny Green}, \bibinfo{person}{Gabriella Barros},
  {and} \bibinfo{person}{Julian Togelius}.} \bibinfo{year}{2019}\natexlab{}.
\newblock \showarticletitle{Intentional computational level design}. In
  \bibinfo{booktitle}{\emph{Proceedings of the Genetic and Evolutionary
  Computation Conference Companion}}. \bibinfo{publisher}{ACM},
  \bibinfo{address}{Prague, Czech Republic}, \bibinfo{pages}{796–803}.
\newblock


\bibitem[\protect\citeauthoryear{Khalifa, Lee, Nealen, and Togelius}{Khalifa
  et~al\mbox{.}}{2018}]%
        {khalifa2018talakat}
\bibfield{author}{\bibinfo{person}{Ahmed Khalifa}, \bibinfo{person}{Scott Lee},
  \bibinfo{person}{Andy Nealen}, {and} \bibinfo{person}{Julian Togelius}.}
  \bibinfo{year}{2018}\natexlab{}.
\newblock \showarticletitle{Talakat: Bullet Hell Generation through Constrained
  Map-Elites}. In \bibinfo{booktitle}{\emph{The Genetic and Evolutionary
  Computation Conference}}. ACM.
\newblock


\bibitem[\protect\citeauthoryear{Kimbrough, Koehler, Lu, and Wood}{Kimbrough
  et~al\mbox{.}}{2008}]%
        {Kimbrough2008Introducing}
\bibfield{author}{\bibinfo{person}{Steven Kimbrough}, \bibinfo{person}{Gary
  Koehler}, \bibinfo{person}{Ming Lu}, {and} \bibinfo{person}{David Wood}.}
  \bibinfo{year}{2008}\natexlab{}.
\newblock \showarticletitle{Introducing a Feasible-Infeasible Two-Population
  (FI-2Pop) Genetic Algorithm for Constrained Optimization: Distance Tracing
  and No Free Lunch}.
\newblock \bibinfo{journal}{\emph{European Journal of Operational Research}}
  \bibinfo{volume}{190} (\bibinfo{date}{10} \bibinfo{year}{2008}),
  \bibinfo{pages}{310--327}.
\newblock
\urldef\tempurl%
\url{https://doi.org/10.1016/j.ejor.2007.06.028}
\showDOI{\tempurl}


\bibitem[\protect\citeauthoryear{Lucas and Volz}{Lucas and Volz}{2019}]%
        {lucas2019tile}
\bibfield{author}{\bibinfo{person}{Simon~M Lucas} {and}
  \bibinfo{person}{Vanessa Volz}.} \bibinfo{year}{2019}\natexlab{}.
\newblock \showarticletitle{Tile pattern KL-divergence for analysing and
  evolving game levels}. In \bibinfo{booktitle}{\emph{Proceedings of the
  Genetic and Evolutionary Computation Conference}}. \bibinfo{pages}{170--178}.
\newblock


\bibitem[\protect\citeauthoryear{McGuinness and Ashlock}{McGuinness and
  Ashlock}{2011}]%
        {mcguinness2011decomposing}
\bibfield{author}{\bibinfo{person}{Cameron McGuinness} {and}
  \bibinfo{person}{Daniel Ashlock}.} \bibinfo{year}{2011}\natexlab{}.
\newblock \showarticletitle{Decomposing the level generation problem with
  tiles}. In \bibinfo{booktitle}{\emph{Congress on Evolutionary Computation}}.
  IEEE, \bibinfo{pages}{849--856}.
\newblock


\bibitem[\protect\citeauthoryear{Perez-Liebana, Samothrakis, Togelius, Schaul,
  and Lucas}{Perez-Liebana et~al\mbox{.}}{2016}]%
        {perez2016general}
\bibfield{author}{\bibinfo{person}{Diego Perez-Liebana},
  \bibinfo{person}{Spyridon Samothrakis}, \bibinfo{person}{Julian Togelius},
  \bibinfo{person}{Tom Schaul}, {and} \bibinfo{person}{Simon~M Lucas}.}
  \bibinfo{year}{2016}\natexlab{}.
\newblock \showarticletitle{General video game ai: Competition, challenges and
  opportunities}. In \bibinfo{booktitle}{\emph{Thirtieth AAAI Conference on
  Artificial Intelligence}}.
\newblock


\bibitem[\protect\citeauthoryear{Persson}{Persson}{2008}]%
        {persson2008infinite}
\bibfield{author}{\bibinfo{person}{Marcus Persson}.}
  \bibinfo{year}{2008}\natexlab{}.
\newblock \showarticletitle{Infinite mario bros}.
\newblock \bibinfo{journal}{\emph{Online Game). Last Accessed: December}}
  \bibinfo{volume}{11} (\bibinfo{year}{2008}).
\newblock


\bibitem[\protect\citeauthoryear{Shaker, Shaker, and Togelius}{Shaker
  et~al\mbox{.}}{2013}]%
        {shaker2013evolving}
\bibfield{author}{\bibinfo{person}{Noor Shaker}, \bibinfo{person}{Mohammad
  Shaker}, {and} \bibinfo{person}{Julian Togelius}.}
  \bibinfo{year}{2013}\natexlab{}.
\newblock \showarticletitle{Evolving Playable Content for Cut the Rope through
  a Simulation-Based Approach.}. In \bibinfo{booktitle}{\emph{AIIDE}}.
\newblock


\bibitem[\protect\citeauthoryear{Shaker, Togelius, Yannakakis, Weber, Shimizu,
  Hashiyama, Sorenson, Pasquier, Mawhorter, Takahashi, et~al\mbox{.}}{Shaker
  et~al\mbox{.}}{2011a}]%
        {shaker20112010}
\bibfield{author}{\bibinfo{person}{Noor Shaker}, \bibinfo{person}{Julian
  Togelius}, \bibinfo{person}{Georgios~N Yannakakis}, \bibinfo{person}{Ben
  Weber}, \bibinfo{person}{Tomoyuki Shimizu}, \bibinfo{person}{Tomonori
  Hashiyama}, \bibinfo{person}{Nathan Sorenson}, \bibinfo{person}{Philippe
  Pasquier}, \bibinfo{person}{Peter Mawhorter}, \bibinfo{person}{Glen
  Takahashi}, {et~al\mbox{.}}} \bibinfo{year}{2011}\natexlab{a}.
\newblock \showarticletitle{The 2010 Mario AI championship: Level generation
  track}.
\newblock \bibinfo{journal}{\emph{Transactions on Computational Intelligence
  and AI in Games}} \bibinfo{volume}{3}, \bibinfo{number}{4}
  (\bibinfo{year}{2011}), \bibinfo{pages}{332--347}.
\newblock


\bibitem[\protect\citeauthoryear{Shaker, Yannakakis, and Togelius}{Shaker
  et~al\mbox{.}}{2011b}]%
        {shaker2011feature}
\bibfield{author}{\bibinfo{person}{Noor Shaker}, \bibinfo{person}{Georgios~N
  Yannakakis}, {and} \bibinfo{person}{Julian Togelius}.}
  \bibinfo{year}{2011}\natexlab{b}.
\newblock \showarticletitle{Feature analysis for modeling game content
  quality}. In \bibinfo{booktitle}{\emph{Computational Intelligence and Games
  Conference}}. IEEE, \bibinfo{pages}{126--133}.
\newblock


\bibitem[\protect\citeauthoryear{Sicart}{Sicart}{2008}]%
        {sicart2008defining}
\bibfield{author}{\bibinfo{person}{Miguel Sicart}.}
  \bibinfo{year}{2008}\natexlab{}.
\newblock \showarticletitle{Defining game mechanics}.
\newblock \bibinfo{journal}{\emph{Game Studies}} \bibinfo{volume}{8},
  \bibinfo{number}{2} (\bibinfo{year}{2008}), \bibinfo{pages}{n}.
\newblock


\bibitem[\protect\citeauthoryear{Smith, Andersen, Mateas, and
  Popovi{\'c}}{Smith et~al\mbox{.}}{2012}]%
        {smith2012case}
\bibfield{author}{\bibinfo{person}{Adam~M Smith}, \bibinfo{person}{Erik
  Andersen}, \bibinfo{person}{Michael Mateas}, {and} \bibinfo{person}{Zoran
  Popovi{\'c}}.} \bibinfo{year}{2012}\natexlab{}.
\newblock \showarticletitle{A case study of expressively constrainable level
  design automation tools for a puzzle game}. In
  \bibinfo{booktitle}{\emph{International Conference on the Foundations of
  Digital Games}}. ACM, \bibinfo{pages}{156--163}.
\newblock


\bibitem[\protect\citeauthoryear{Smith and Mateas}{Smith and Mateas}{2011}]%
        {smith2011answer}
\bibfield{author}{\bibinfo{person}{Adam~M Smith} {and} \bibinfo{person}{Michael
  Mateas}.} \bibinfo{year}{2011}\natexlab{}.
\newblock \showarticletitle{Answer set programming for procedural content
  generation: A design space approach}.
\newblock \bibinfo{journal}{\emph{IEEE Transactions on Computational
  Intelligence and AI in Games}} \bibinfo{volume}{3}, \bibinfo{number}{3}
  (\bibinfo{year}{2011}), \bibinfo{pages}{187--200}.
\newblock


\bibitem[\protect\citeauthoryear{Smith, Whitehead, Mateas, Treanor, March, and
  Cha}{Smith et~al\mbox{.}}{2011}]%
        {smith2011launchpad}
\bibfield{author}{\bibinfo{person}{Gillian Smith}, \bibinfo{person}{Jim
  Whitehead}, \bibinfo{person}{Michael Mateas}, \bibinfo{person}{Mike Treanor},
  \bibinfo{person}{Jameka March}, {and} \bibinfo{person}{Mee Cha}.}
  \bibinfo{year}{2011}\natexlab{}.
\newblock \showarticletitle{Launchpad: A rhythm-based level generator for 2-d
  platformers}.
\newblock \bibinfo{journal}{\emph{Transactions on computational intelligence
  and AI in games}} \bibinfo{volume}{3}, \bibinfo{number}{1},
  \bibinfo{pages}{1--16}.
\newblock


\bibitem[\protect\citeauthoryear{Sorenson and Pasquier}{Sorenson and
  Pasquier}{2010}]%
        {sorenson2010towards}
\bibfield{author}{\bibinfo{person}{Nathan Sorenson} {and}
  \bibinfo{person}{Philippe Pasquier}.} \bibinfo{year}{2010}\natexlab{}.
\newblock \showarticletitle{Towards a generic framework for automated video
  game level creation}. In \bibinfo{booktitle}{\emph{European Conference on the
  Applications of Evolutionary Computation}}. Springer,
  \bibinfo{pages}{131--140}.
\newblock


\bibitem[\protect\citeauthoryear{Summerville, Snodgrass, Guzdial, Holmg{\aa}rd,
  Hoover, Isaksen, Nealen, and Togelius}{Summerville et~al\mbox{.}}{2018}]%
        {summerville2018procedural}
\bibfield{author}{\bibinfo{person}{Adam Summerville}, \bibinfo{person}{Sam
  Snodgrass}, \bibinfo{person}{Matthew Guzdial}, \bibinfo{person}{Christoffer
  Holmg{\aa}rd}, \bibinfo{person}{Amy~K Hoover}, \bibinfo{person}{Aaron
  Isaksen}, \bibinfo{person}{Andy Nealen}, {and} \bibinfo{person}{Julian
  Togelius}.} \bibinfo{year}{2018}\natexlab{}.
\newblock \showarticletitle{Procedural content generation via machine learning
  (pcgml)}.
\newblock \bibinfo{journal}{\emph{IEEE Transactions on Games}}
  \bibinfo{volume}{10}, \bibinfo{number}{3} (\bibinfo{year}{2018}),
  \bibinfo{pages}{257--270}.
\newblock


\bibitem[\protect\citeauthoryear{Togelius and Dahlskog}{Togelius and
  Dahlskog}{2013}]%
        {togelius2013patterns}
\bibfield{author}{\bibinfo{person}{Julian Togelius} {and}
  \bibinfo{person}{Steve Dahlskog}.} \bibinfo{year}{2013}\natexlab{}.
\newblock \showarticletitle{Patterns as objectives for level generation}. In
  \bibinfo{booktitle}{\emph{Proceedings of the Second Workshop on Design
  Patterns in Games;}}. ACM.
\newblock


\bibitem[\protect\citeauthoryear{Togelius, Karakovskiy, and
  Baumgarten}{Togelius et~al\mbox{.}}{2010a}]%
        {Togelius2009The}
\bibfield{author}{\bibinfo{person}{Julian Togelius}, \bibinfo{person}{Sergey
  Karakovskiy}, {and} \bibinfo{person}{Robin Baumgarten}.}
  \bibinfo{year}{2010}\natexlab{a}.
\newblock \showarticletitle{The 2009 Mario AI Competition}. In
  \bibinfo{booktitle}{\emph{2010 IEEE World Congress on Computational
  Intelligence, WCCI 2010 - 2010 IEEE Congress on Evolutionary Computation, CEC
  2010}}.
\newblock
\showISBNx{9781424469109}


\bibitem[\protect\citeauthoryear{Togelius, Karakovskiy, and
  Baumgarten}{Togelius et~al\mbox{.}}{2010b}]%
        {togelius20102009}
\bibfield{author}{\bibinfo{person}{Julian Togelius}, \bibinfo{person}{Sergey
  Karakovskiy}, {and} \bibinfo{person}{Robin Baumgarten}.}
  \bibinfo{year}{2010}\natexlab{b}.
\newblock \showarticletitle{The 2009 mario ai competition}. In
  \bibinfo{booktitle}{\emph{IEEE Congress on Evolutionary Computation}}. IEEE,
  \bibinfo{pages}{1--8}.
\newblock


\bibitem[\protect\citeauthoryear{Togelius, Shaker, Karakovskiy, and
  Yannakakis}{Togelius et~al\mbox{.}}{2013}]%
        {togelius2013mario}
\bibfield{author}{\bibinfo{person}{Julian Togelius}, \bibinfo{person}{Noor
  Shaker}, \bibinfo{person}{Sergey Karakovskiy}, {and}
  \bibinfo{person}{Georgios~N Yannakakis}.} \bibinfo{year}{2013}\natexlab{}.
\newblock \showarticletitle{The mario ai championship 2009-2012}.
\newblock \bibinfo{journal}{\emph{AI Magazine}} \bibinfo{volume}{34},
  \bibinfo{number}{3} (\bibinfo{year}{2013}), \bibinfo{pages}{89--92}.
\newblock


\bibitem[\protect\citeauthoryear{Togelius, Yannakakis, Stanley, and
  Browne}{Togelius et~al\mbox{.}}{2011}]%
        {togelius2011search}
\bibfield{author}{\bibinfo{person}{Julian Togelius},
  \bibinfo{person}{Georgios~N Yannakakis}, \bibinfo{person}{Kenneth~O Stanley},
  {and} \bibinfo{person}{Cameron Browne}.} \bibinfo{year}{2011}\natexlab{}.
\newblock \showarticletitle{Search-based procedural content generation: A
  taxonomy and survey}.
\newblock \bibinfo{journal}{\emph{IEEE Transactions on Computational
  Intelligence and AI in Games}} \bibinfo{volume}{3}, \bibinfo{number}{3}
  (\bibinfo{year}{2011}), \bibinfo{pages}{172--186}.
\newblock


\bibitem[\protect\citeauthoryear{Yannakakis and Togelius}{Yannakakis and
  Togelius}{2011}]%
        {yannakakis2011experience}
\bibfield{author}{\bibinfo{person}{Georgios~N Yannakakis} {and}
  \bibinfo{person}{Julian Togelius}.} \bibinfo{year}{2011}\natexlab{}.
\newblock \showarticletitle{Experience-driven procedural content generation}.
\newblock \bibinfo{journal}{\emph{IEEE Transactions on Affective Computing}}
  \bibinfo{volume}{2}, \bibinfo{number}{3} (\bibinfo{year}{2011}),
  \bibinfo{pages}{147--161}.
\newblock


\end{thebibliography}

\end{document}